\documentclass[10pt,twocolumn,letterpaper]{article}

\usepackage[pagenumbers]{cvpr} 
\usepackage{soul}
\usepackage[table]{xcolor} 
\usepackage{changepage,threeparttable} 

\usepackage{ifsym}
\usepackage{marvosym}
\usepackage{graphicx}
\usepackage{amsmath}
\usepackage{amssymb}
\usepackage{booktabs}
\usepackage{multirow}
\usepackage[utf8]{inputenc}

\usepackage[super]{nth}
\DeclareUnicodeCharacter{2212}{−}

\usepackage[accsupp]{axessibility}  

\usepackage[pagebackref,breaklinks,colorlinks]{hyperref}

\usepackage[capitalize]{cleveref}
\crefname{section}{Sec.}{Secs.}
\Crefname{section}{Section}{Sections}
\Crefname{table}{Table}{Tables}
\crefname{table}{Tab.}{Tabs.}

\begin{document}

\title{Joint Token Pruning and Squeezing Towards More Aggressive Compression of Vision Transformers}

\author{
    Siyuan Wei$^1$\thanks{The first two authors contributed equally to this work} \qquad
    Tianzhu Ye$^2$\footnotemark[1] \qquad
    Shen Zhang$^1$ \qquad
    Yao Tang$^1$ \qquad
    Jiajun Liang$^1$\thanks{Corresponding author} \\
    $^1$MEGVII Technology \qquad
    $^2$Tsinghua University \\
    {\tt\small \{weisiyuan, zhangshen,tangyao02,liangjiajun\}@megvii.com,ytz20@mails.tsinghua.edu.cn}
}
\maketitle

\begin{abstract}

Although vision transformers (ViTs) have shown promising results in various computer vision tasks recently, their high computational cost limits their practical applications. Previous approaches that prune redundant tokens have demonstrated a good trade-off between performance and computation costs. Nevertheless, errors caused by pruning strategies can lead to significant information loss. Our quantitative experiments reveal that the impact of pruned tokens on performance should be noticeable. To address this issue, we propose a novel joint Token Pruning \& Squeezing module (TPS) for compressing vision transformers with higher efficiency. Firstly, TPS adopts pruning to get the reserved and pruned subsets. Secondly, TPS squeezes the information of pruned tokens into partial reserved tokens via the unidirectional nearest-neighbor matching and similarity-based fusing steps. Compared to state-of-the-art methods, our approach outperforms them under all token pruning intensities. Especially while shrinking DeiT-tiny\&small computational budgets to 35\%, it improves the accuracy by 1\%-6\% compared with baselines on ImageNet classification. The proposed method can accelerate the throughput of DeiT-small beyond DeiT-tiny, while its accuracy surpasses DeiT-tiny by 4.78\%. Experiments on various transformers demonstrate the effectiveness of our method, while analysis experiments prove our higher robustness to the errors of the token pruning policy. Code is available at \url{https://github.com/megvii-research/TPS-CVPR2023}.
\end{abstract}

\section{Introduction}



The transformer architecture has become popular for various natural language processing (NLP) tasks, and its improved variants have been adopted for many vision tasks. Vision transformers (ViTs) \cite{dosovitskiy2020image} leverage the long-range dependencies of self-attention mechanisms to achieve excellent performance, often surpassing that of CNNs. In addition to the vanilla ViT architecture, recent studies \cite{liu2021swin,wang2021pyramid,wu2021cvt} have explored hybrid ViT designs incorporating convolution layers and multi-scale architectures. Despite their excellent performance, transformers still require relatively high computational budgets. This is due to the quadratic computation and memory costs associated with token length. To address this issue, contemporary approaches \cite{rao2021dynamicvit,tang2022patch,yin2022vit,fayyazadaptive,liang2022not,xu2022evo,kong2022spvit,pan2021ia} propose pruning redundant tokens. They trade acceptable performance degradation for a more cost-effective model. Knowledge distillation  \cite{hinton2015distilling} and other techniques can further mitigate the resulting performance drop.

\begin{figure}[t]
  \centering
  \includegraphics[width=0.9\linewidth]{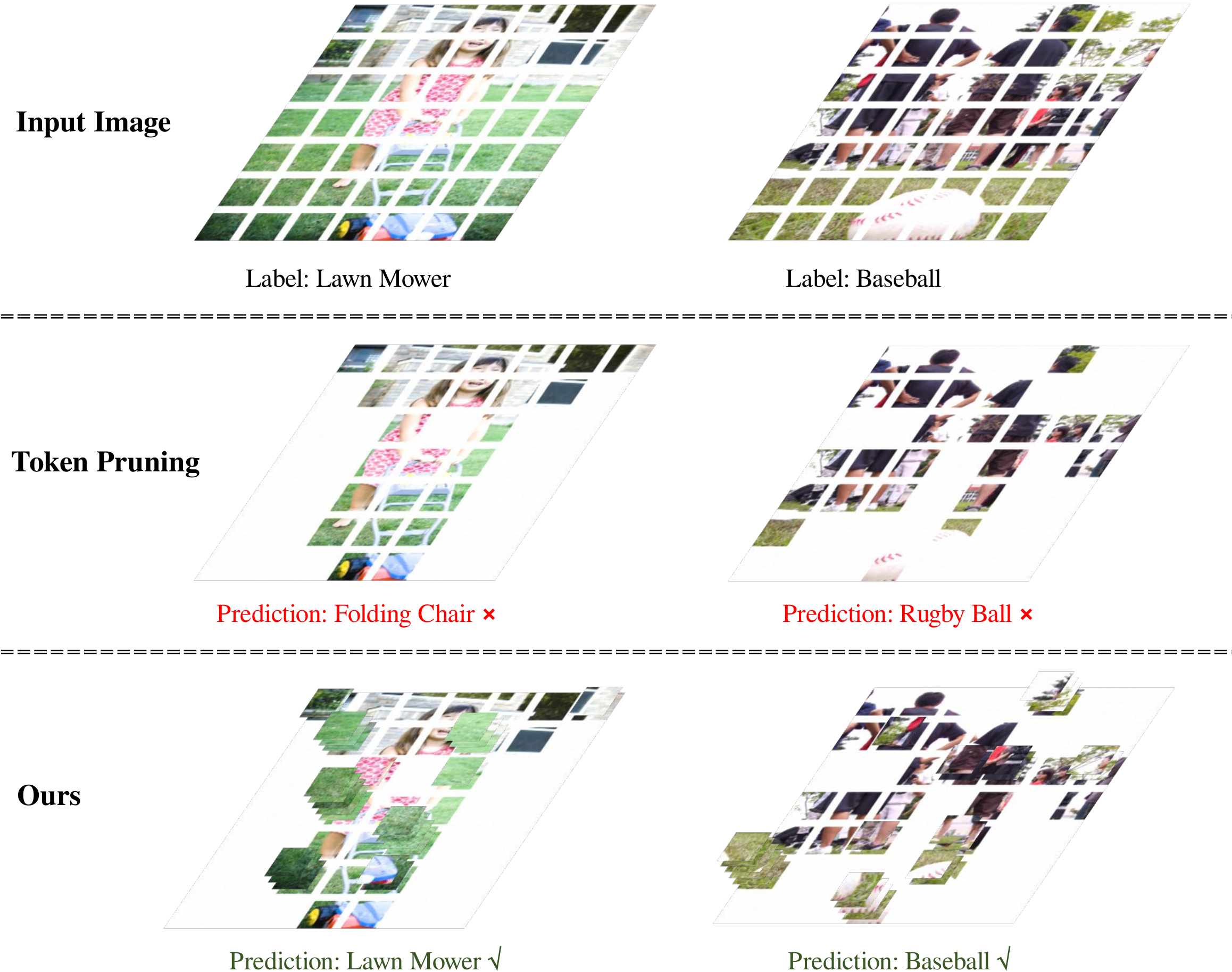}
  \caption{Comparisons between token pruning paradigm ~\cite{rao2021dynamicvit} (the 2nd row) and our joint Token Pruning \& Squeezing (the 3rd row). The context information, such as the sod in the examples, is helpful for prediction but is discarded. Our method remits the information loss by squeezing the pruned tokens into reserved ones instead of naively dropping them, as indicated by the stacked patches. By this design, we could apply more aggressive token pruning with less performance drop. The example results are from the ImageNet1K~\cite{deng2009imagenet}, and we reduce the actual patches grid $14\times14$ to $7\times7$ for visualization clarity.}
  \label{fig:cases}
\end{figure}

\begin{figure*}
    \centering
    \includegraphics[width=1.0\linewidth]{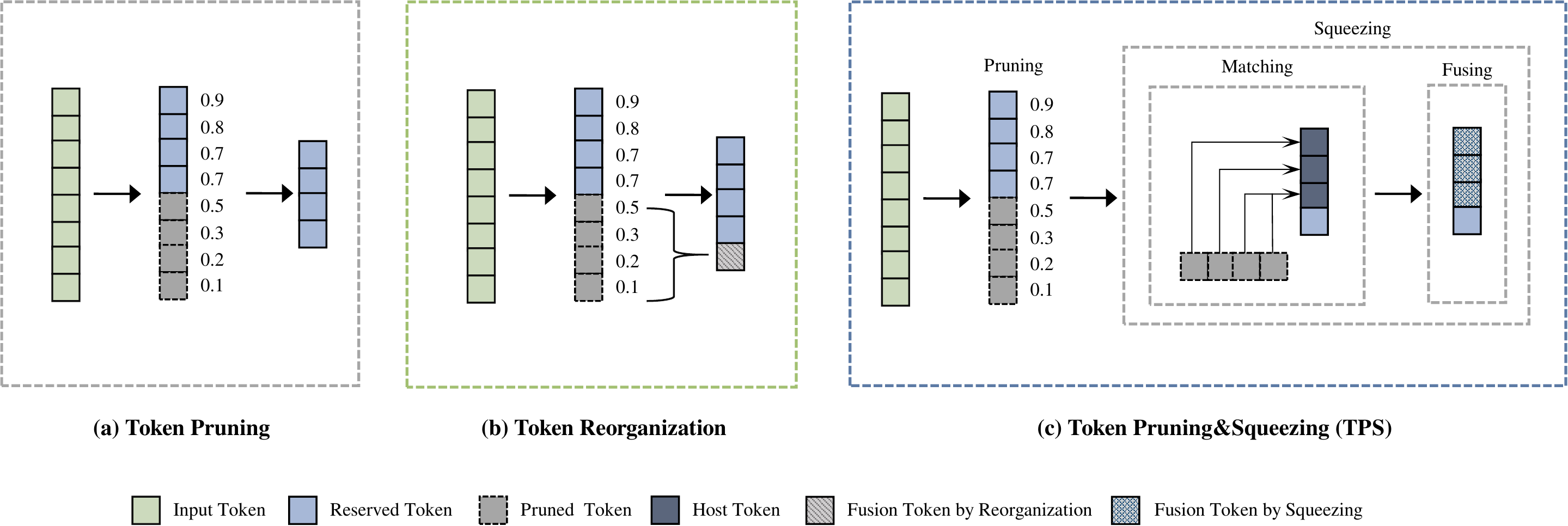}
    \caption{Comparison among token pruning \cite{rao2021dynamicvit}, token reorganization \cite{liang2022not}, and our Token Pruning \& Squeezing (TPS). Token pruning discards the pruned tokens; token reorganization aggregates pruned tokens into one without considering the discrepancy among them. To shrink tokens more efficiently, our TPS excavates the host token hiding in the reserved subset and squeezes similar pruned tokens into corresponding host tokens. }
    \label{fig:arch-a}
\end{figure*}

However, a steep drop in performance is inevitable as pruning tokens further increases because both essential subject and auxiliary context information drop significantly, especially when the number of reserved tokens is closely below 10. Aggressive token pruning could lead to incomplete subject and background context loss, causing the wrong prediction, as shown in \cref{fig:cases}. Specifically, the background tokens containing sod help recognize the input image as a lawn mower rather than a folding chair. Meanwhile, missing subject tokens make the baseball indistinguishable from a rugby ball. To regain adequate information from pruned tokens, EViT~\cite{liang2022not} and Evo-ViT~\cite{xu2022evo} propose aggregating pruned tokens as one, as shown in \cref{fig:arch-a} (b). Still, they neglect the discrepancy among these tokens, leading to feature collapse and hindering more aggressive token pruning.

Towards more aggressive pruning, we argue that information in pruned tokens deserves better treatment. We did a toy experiment to answer what accuracy token pruning could achieve if it applied the reversed pruning policy in the first pruned transformer block as \cref{fig:reversed_cases} shows. Taking dynamicViT~\cite{rao2021dynamicvit} as a case study, the performance of reversed policy is enough to bring extra accuracy complementary to the original one (denoted by bonus accuracy). Moreover, this phenomenon would become more significant as pruning continues (red line in \cref{fig:reversed_cases}.).

To conserve the information from the pruned tokens, we propose a Joint Token Pruning \& Squeezing (TPS) module to accommodate more aggressive compression of ViTs. TPS module utilizes a feature dispatch mechanism that squeezes essential features from pruned tokens into reserved ones, as shown in \cref{fig:arch-a} (c). Firstly, based on the scoring result, the TPS module divides input tokens into two complementary subsets: the reserved and pruned sets. Secondly, instead of discarding or collapsing tokens from the pruned set into a single one, we employ a unidirectional nearest-neighbor matching algorithm to dispatch each of them independently to the associated reserved token dubbed as the host token. This design reduces information loss without sacrificing computational efficiency. Subsequently, we apply a similarity-based fusing way to squeeze the features of matched pruned tokens into corresponding host tokens while the non-selected reserved tokens remain identical. This design reduces the context information loss while retaining a reasonable computation budget. We can easily achieve hardware-friendly constant shape inference when fixing the cardinality of the reserved token set. Furthermore, we introduce two flexible variants: the inter-block version dTPS and the intra-block version eTPS, which are essentially plug-and-play blocks for both vanilla ViTs and hybrid ViTs.

We conduct extensive experiments on two datasets: ImageNet1K~\cite{deng2009imagenet} and large fine-grained dataset iNaturalist 2019~\cite{fgvc6} to prove our efficiency, flexibility, and robustness. Firstly, experiments under different token pruning settings demonstrate the superior performance of our TPS while operating more aggressive compression compared with token pruning~\cite{rao2021dynamicvit} and token reorganization~\cite{liang2022not}; further comparisons with state-of-the-art transformers  ~\cite{touvron2021training,jiang2021all,yue2021vision,wang2021pyramid,zeng2022not,meng2022adavit,fayyazadaptive,yin2022vit} show our promising efficiency. Secondly, we manifest the flexibility of our TPS by integrating it into popular ViTs, including both vanilla ViTs and hybrid ViTs. Finally, the evaluations under the random token selection policy confirm the higher robustness of our TPS.

Overall, our contributions are summarized as follows:
\begin{itemize}
    \item We propose the joint Token Pruning \& Squeezing (TPS) and its two variants: dTPS and eTPS, to conserve the information of discarded tokens and facilitate more aggressive compression of vision transformers.
    \item Extensive experiments demonstrate our higher performance compared with prior approaches. Especially while compressing GFLOPs of DeiT-small\&tiny to 35\%, our TPS outperforms baselines with accuracy improvements of 1\%-6\%. 
    \item Broadest experiments applying our method to vanilla ViTs and hybrid ViTs show our flexibility, while the analysis experiments prove that our TPS is more robust than token pruning and token reorganization.
\end{itemize}

\section{Related Work}
Since the transformer~\cite{vaswani2017attention} was proved efficacious in NLP tasks, numerous studies have explored methods to acclimate the transformer architecture to computer vision tasks~\cite{dosovitskiy2020image,el2021training,fan2021multiscale,carion2020end,cheng2021per,liu2021swin,he2021transreid,mao2021voxel,pan20213d,rao2021global,yu2021pointr,ranftl2021vision,wang2021end}, including vanilla ViTs and hybrid ViTs. 

\textbf{Vanilla ViTs.} Following the ``primary ViT", a series of vision transformer variants inherit the central architecture and evolve from diverse perspectives, which we call the vanilla ViTs in this paper. DeiT~\cite{touvron2021training} surpasses standard CNNs and ViT by introducing a distillation token to learn from a teacher network. LV-ViT~\cite{jiang2021all} presents a new training objective called token labeling. T2T-ViT~\cite{yuan2021tokens} recursively aggregates neighboring tokens into one token, while PS-ViT~\cite{yue2021vision} introduces a progressive sampling module that selects informative tokens.

\textbf{Hybrid ViTs.} Besides, recent studies~\cite{wang2021pyramid,wu2021cvt,li2021localvit}  incorporate convolutional layers and employ multi-scale architectures to lower the cost of computations and memory, which we call the hybrid ViTs. Swin Transformer~\cite{liu2021swin} modified ViT with the multi-stage architecture and shifted window-based self-attention. CVT~\cite{wu2021cvt} presents a hierarchical architecture facilitated by the convolutional token embedding layer. PVT~\cite{wang2021pyramid} introduces the pyramid architecture of the transformer and develops the spatial-reduction attention (SRA) to reduce the cost further.

\textbf{Token Pruning.} Considering the spatial redundancy of input images,  many researchers aim at discarding nonessential tokens with an acceptable performance drop. Tang \etal~\cite{tang2022patch} propose to approximate the impact of patches and discard inattentive patches in a top-down paradigm. DynamicViT~\cite{rao2021dynamicvit} and AdaViT~\cite{meng2022adavit} employ the learnable heads to score tokens and discard less informative ones with a fixed pruning ratio. A-ViT~\cite{yin2022vit} and ATS~\cite{fayyazadaptive} go further by sampling tokens with an input-dependent number. However, mainstream deep learning frameworks do not strongly support dynamic token length inference. The main disadvantage of token pruning models is the pruned information loss which leads to a drop in accuracy and limits more aggressive token pruning. To tackle this, Evo-ViT~\cite{xu2022evo}, EViT~\cite{liang2022not}, and SPViT~\cite{kong2022spvit} preserve the background context by collapsing the pruned tokens into one token reorganization, which is called token reorganization. Token reorganization remits the pruned token information loss, but a noticeable performance drop can still be observed, especially regarding a higher pruning ratio of tokens. Furthermore, relevant auxiliary strategies are proposed to facilitate token pruning. SPViT~\cite{kong2022spvit} employs a layer-to-phase progressive training strategy, while IA-RED$^2$  performs a hierarchical training scheme. The 
complicated training schemes help improve performances but also draw into more hyper-parameters and optimization difficulties.

We investigate the drawbacks of current token pruning methods and invent a novel token reduction approach: joint token Pruning \& Squeezing with higher efficiency,  robustness, and flexibility, which only requires fine-tuning pre-trained models easily.


\section{Method}

\begin{figure}[t]
    \centering
    \includegraphics[width=0.8\linewidth]{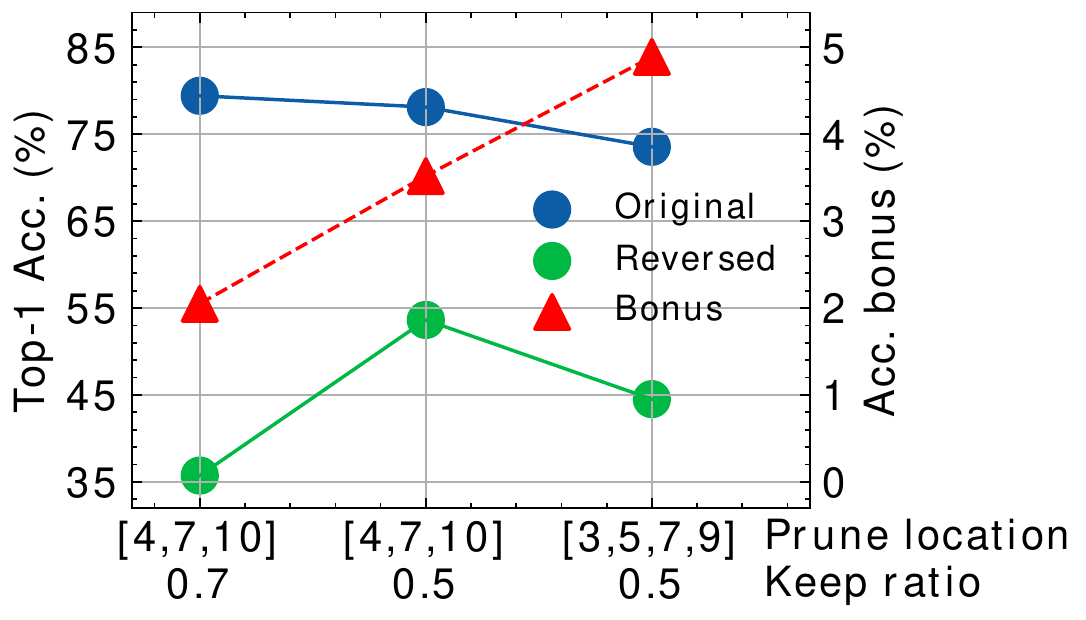}
    \caption{ImageNet1K results of DynamicViT~\cite{rao2021dynamicvit} on DeiT-small~\cite{touvron2021training} under the original policy and reversed policy. The reversed policy means exchanging reserved and pruned tokens in the first pruned layer. The right vertical axis implies the bonus accuracy dedicated by the cases that only the reversed policy predicts rightly.}
    \label{fig:reversed_cases}
\end{figure}

\begin{figure*}
    
    \centering
    \begin{subfigure}{0.4\linewidth}
        \centering
        \includegraphics[width=1.0\linewidth]{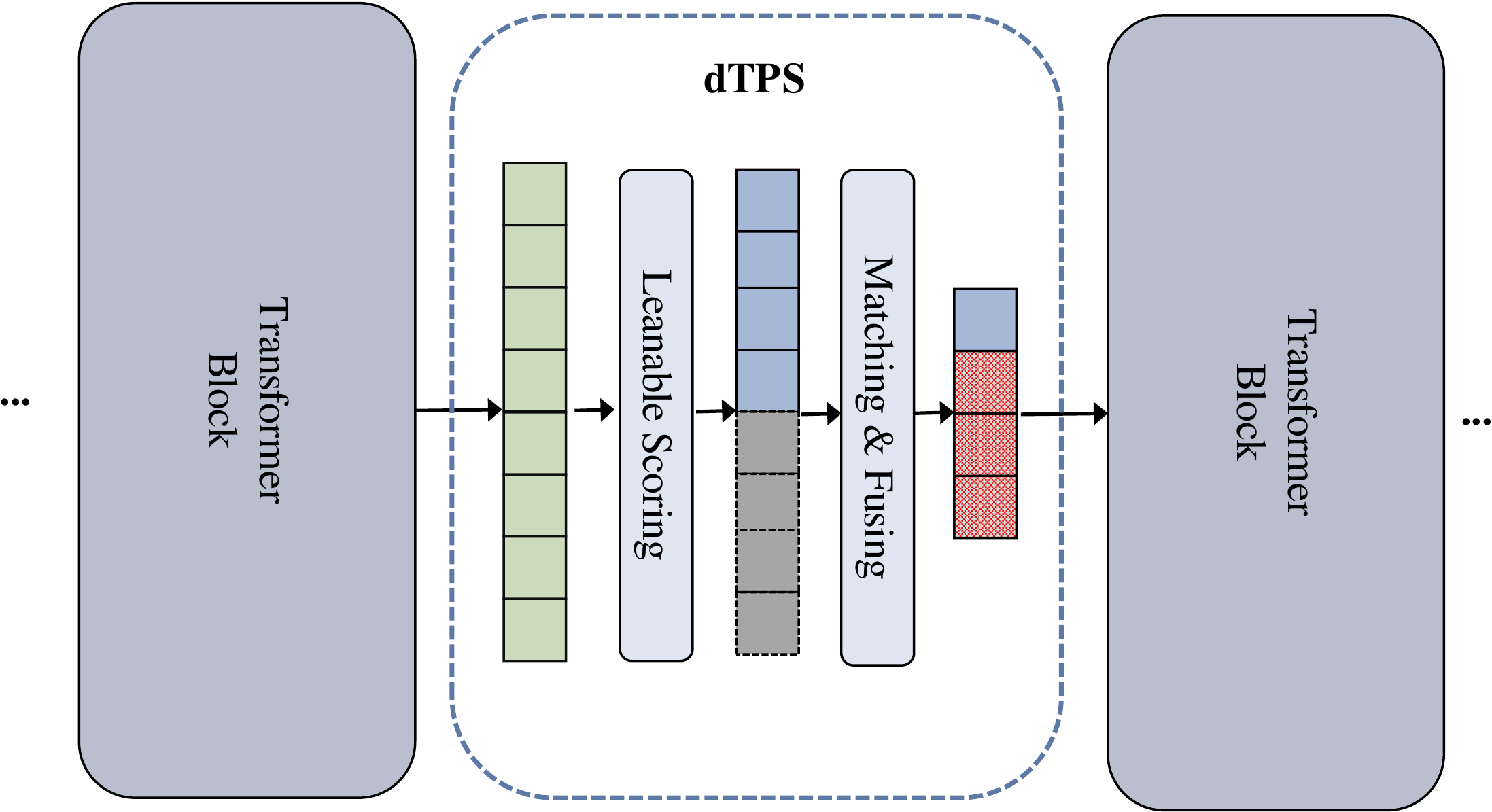}
        \caption{The inter-block variant of our TPS: dTPS.}
        \label{fig:arch-b}
    \end{subfigure}
    \hspace{10mm}
    \begin{subfigure}{0.4\linewidth}
        \centering
        \includegraphics[width=1.0\linewidth]{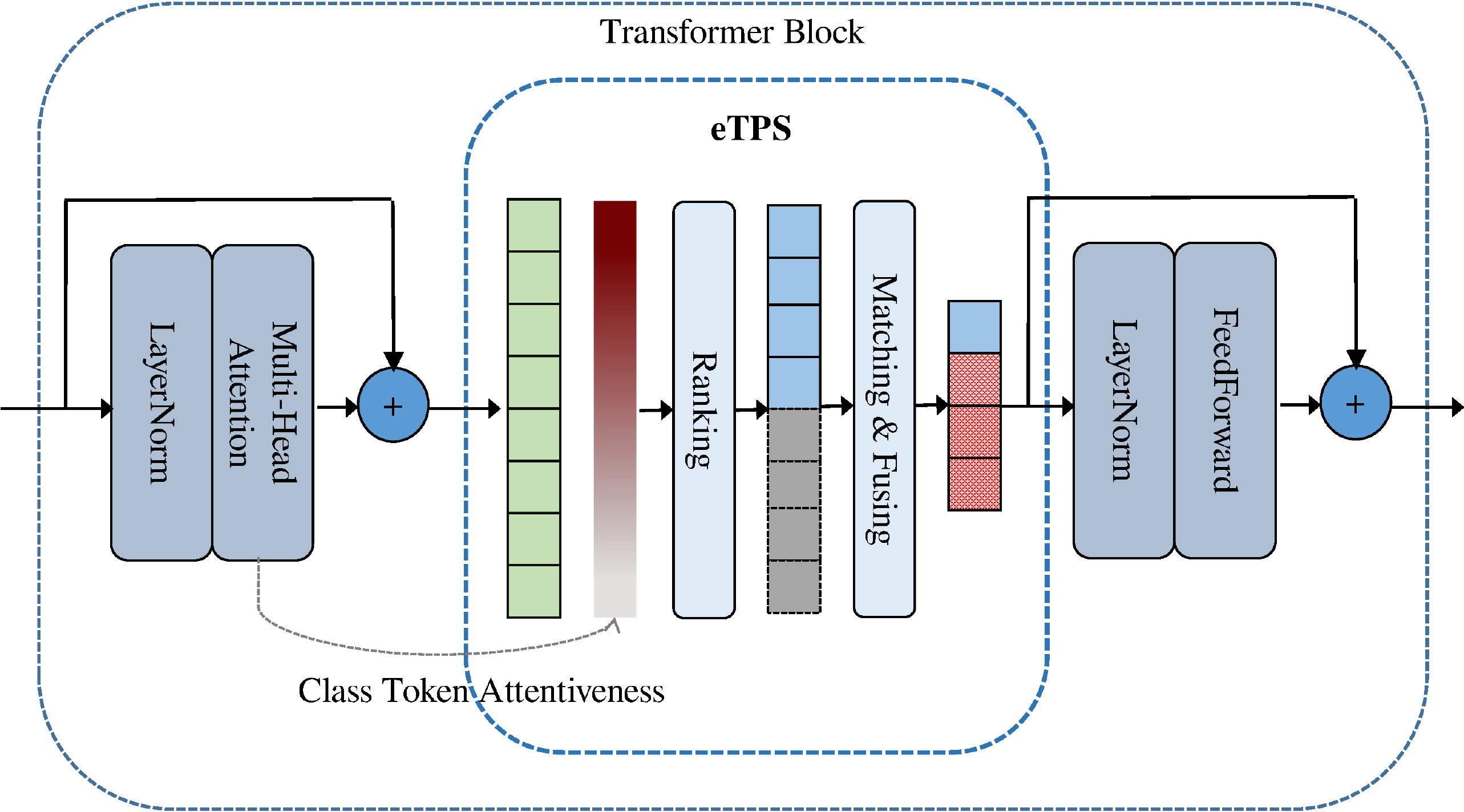}
        \caption{The intra-block variant of our TPS: eTPS.}
        \label{fig:arch-c}
    \end{subfigure}
        
    \caption{Two variants of our TPS. TPS can be plugged in both the inter-block and intra-block ways. For a fair comparison with DynamicViT~\cite{rao2021dynamicvit} and EViT \cite{liang2022not}, our dTPS and eTPS adopt the same token scoring methods as the two baselines. The legend of tokens is the same as the Fig.~\ref{fig:arch-a}.}
    \label{fig:arch}
\end{figure*}
  
\subsection{Motivation}
\label{motivation}

 To quantitatively verify the discarded information of pruned tokens, we conduct a toy experiment on DynamicViT~\cite{rao2021dynamicvit} as Fig.~\ref{fig:reversed_cases} shows. It is easy to agree that the performance of pruned model declines as the pruning becomes more aggressive. Nevertheless, by exchanging reserved and pruned tokens (dubbed as the reversed policy in Fig.~\ref{fig:reversed_cases}), we find that the pruned tokens can still handle partial cases correctly. Furthermore, the bonus accuracy, which is dedicated by the cases that only the reversed policy predicts rightly, rises along with the token pruning intensity. It implies that the exclusive information from pruned tokens matters more while the token pruning intensity grows.
 
 These fun facts motivate us to assimilate the pruned tokens into the reserved tokens to prevent information loss, as shown in Fig.~\ref{fig:cases}. As shown in Fig.~\ref{fig:arch-a} (c), TPS employs two steps to compress ViTs, including token pruning and squeezing.

\subsection{Token Pruning} \label{token_score_and_pruning}
    
    In this section, we briefly review the basic procedure of token pruning. Note that our TPS is compatible with any token pruning techniques. Here, we introduce two variants of TPS: dTPS and eTPS, to cover both intra-block and inter-block token compression shown in \cref{fig:arch}. They follow the pruning parts of two baselines for a fair comparison with two typical baselines~\cite{rao2021dynamicvit,liang2022not}.
    


    As shown in \cref{fig:arch}, dTPS adopts the learnable token score prediction head from dynamicViT~\cite{rao2021dynamicvit} and samples the binary decision mask by Straight-Through Gumbel-Softmax~\cite{jang2016categorical} for differentiability; eTPS utilizes the class token attention values to measure tokens' importance as EViT~\cite{liang2022not}. In the inference stage of both variants, based on token scores, we devise the token selection policy using the Top-k operation with a fixed given token reduction ratio ${\rho}$. Both variants ensure the constant shape to benefit from the inference optimization on the computation graph. The tokens are separated into two subsets, $S^r$ and $S^p$, where the reserved tokens are placed in $S^r$ and the pruned ones are placed in $S^p$. More implementation details can be found in our codes.

    
\subsection{Token Squeezing}

    After reserved \& pruned tokens are split, we introduce our token squeezing part. Considering that the reserved ones contribute the majority of correct predictions, we aim to design a procedure that retains most of the attentive tokens while compressing information from rest, preserving the model's overall performance. To avoid generating extra tokens as ~\cite{liang2022not,kong2022spvit}, we inject pruned tokens into similar reserved tokens. So, we apply a unidirectional nearest-neighbor matching algorithm from $S^p$ to $S^r$ in a many-to-one manner. After that, we employ a similarity-based fusing method to assimilate information from pruned tokens into partial reserved tokens. We summarize the above process as two steps: \textit{matching} and \textit{fusing}.

    \textbf{Matching}. Given the two subsets $S^r$ and $S^p$, $I^r$ and $I^p$ are the corresponding token indices of $S^r$ and $S^p$. A similarity matrix $c_{i,j}$ for all $i \in I^p$ and $j \in I^r$ represents the interactions between the tokens for matching. For each pruned token $\boldsymbol{x}_i \in S^p$, we find its nearest token $\boldsymbol{x}_*^{host} \in S^r$ from the reserved token set $S^r$  as its \textit{host token}: 
    \begin{equation}\label{argmax}
        \boldsymbol{x}_*^{host} =\mathop{argmax}\limits_{\boldsymbol{x}_j \in S^r}{c_{i,j}}.
    \end{equation}
    Note that since the token matching step is unidirectional from $S^p$ to $S^r$, multiple pruned tokens can share the same host token and not each reserved token can serve as a host token. We then record the matching results in a mask matrix $\boldsymbol{M} \in \mathbb{R}^{N^p \times N^r}$ and its values are decided by:
    \begin{equation}
        m_{i,j}=\begin{cases}
        1, & \boldsymbol{x}_j\ is\ the\ host\ token\ of\ \boldsymbol{x}_i, \\
        0, & otherwise,
        \end{cases}
    \end{equation}
    where $N^p$ and $N^r$ denote the token number of two subsets. The mask helps that the following fusing step can be conducted with regular matrix operations on $S^r$ and $S^p$ while excluding the influence from non-matched pairs.

    Although the attention map is a natural and free choice to measure interactions among tokens, we can acquire higher performances with the cosine similarity between $S^r$ and $S^p$ as the ablation experiment in Section.~\ref{ablation_study} discusses. Therefore in all of our experiments, the similarity matrix is defined as:
    \begin{equation}
        c_{i,j} = \frac{{\boldsymbol{x}_i^T}{\boldsymbol{x}_j}}{\|\boldsymbol{x}_i\|\|\boldsymbol{x}_j\|}, for\ i \in I^p, j \in I^r.
    \end{equation}
    Since the similarity matrix $c_{i,j}$ is generated directly from input features, no extra parameters are introduced in the matching step.
    
    \textbf{Fusing}. Simply averaging tokens can lead to feature dispersion because of discrepancies among the different tokens. EViT~\cite{liang2022not} utilizes the token importance scores to re-weight the aggregated tokens. Separately, we use a similarity-based weighting scheme. It expands the influence of closer tokens to the host tokens while also avoiding potential flaws from imperfect token scoring. As previously mentioned, the fusing step encompasses all tokens from two subsets and is controlled by the mask $\boldsymbol{M}$ to ensure that only host tokens and pruned tokens are mixed. This introduces a few redundant computations but increases practical training \& inference throughput due to the efficiency of regular matrix operations.

    Specifically, the reserved token $\boldsymbol{x}_j$ is updated by fusing the original feature and pruned tokens' features as follows:
    \begin{equation} \label{fusing}
        \boldsymbol{y}_j={w}_j \boldsymbol{x}_j + \sum_{\boldsymbol{x}_i \in S^p}{w_i \boldsymbol{x}_i},
    \end{equation}
    where $w_i$ is the weight of each pruned token $\boldsymbol{x}_i \in S^p$, ${w}_j$ is the weight of the reserved token itself, and $\boldsymbol{y}_j$ is the updated one. The fusing weight $w_i$ depends on the mask value $m_{i,j}$ and similarity $c_{i,j}$:
    \begin{equation}
        w_i=\frac{\exp(c_{i,j})m_{i,j}}{ \sum_{\boldsymbol{x}_i \in S^p}\exp(c_{i,j})m_{i,j}+\mathrm{e} }.
    \end{equation}
    The reserved token always has the largest fusing weight $w_j$, as the similarity between ${x}_j$ and itself equals to $1$:
    \begin{equation}
        w_j=\frac{\mathrm{e}}{ \sum_{\boldsymbol{x}_i \in S^p}\exp(c_{i,j})m_{i,j}+\mathrm{e} }.
    \end{equation}
    According to the above equations, the reserved tokens that have not been chosen as host tokens remain unchanged, while the pruned tokens are squeezed into host tokens and replace the original ones. 
    
    As can be seen, our matching and fusing steps ensure that the number of processed tokens equals the number of reserved tokens, thereby maintaining a constant shape for efficient inference.

\subsection{TPS on Hybrid ViTs} 
    To prove our flexibility and generalization across different transformers, we also conduct experiments in hybrid ViTs~\cite{wang2021pyramid,wu2021cvt}. For plain transformer blocks, our TPS modules can be easily inserted to reduce the token number and achieve a significant speedup. If the layer contains operations that require a complete spatial structured input: e.g., convolution or pooling, the operation of our TPS will be adjusted slightly. For example, in PVT ~\cite{wang2021pyramid} models, the TPS module is inserted before the first block of each stage with token pruning applied and generates the decision policy $\boldsymbol{D}$. For the attention layer, we decrease the token dimension size of the input and consequent query $\boldsymbol{Q}$. If the spatial-reduction layer is employed inside, the dropped token features are complemented with zeros to maintain the structured spatial input. More details can be found in supplementary materials.

\section{Experiment}

\begin{figure*}
    \centering
    \begin{subfigure}{0.4\linewidth}
        \includegraphics[width=1.0\linewidth]{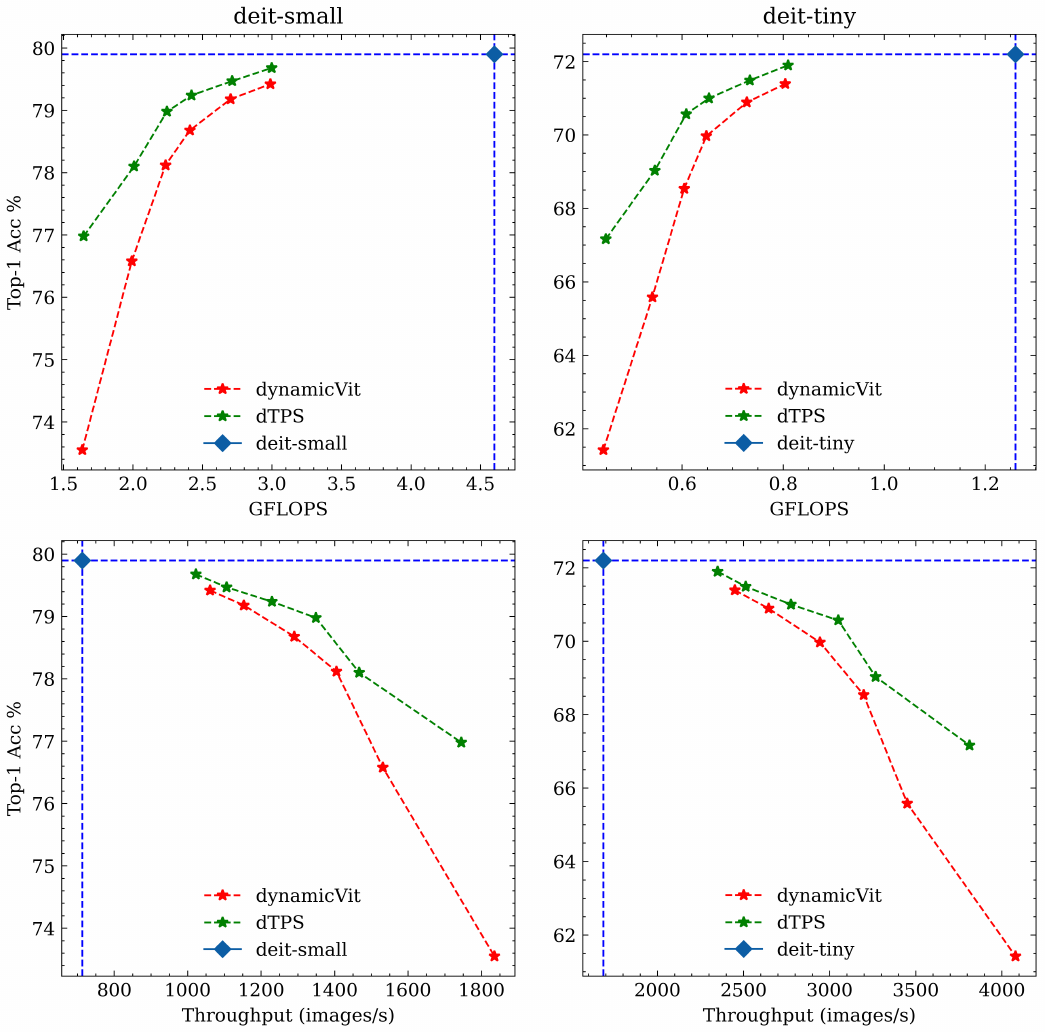}
        \caption{Comparison between our dTPS and dynamicViT on DeiT.}
        \label{fig:dyn30-a}
    \end{subfigure}
    \hspace{10mm}
    \begin{subfigure}{0.4\linewidth}
        \includegraphics[width=1.0\linewidth]{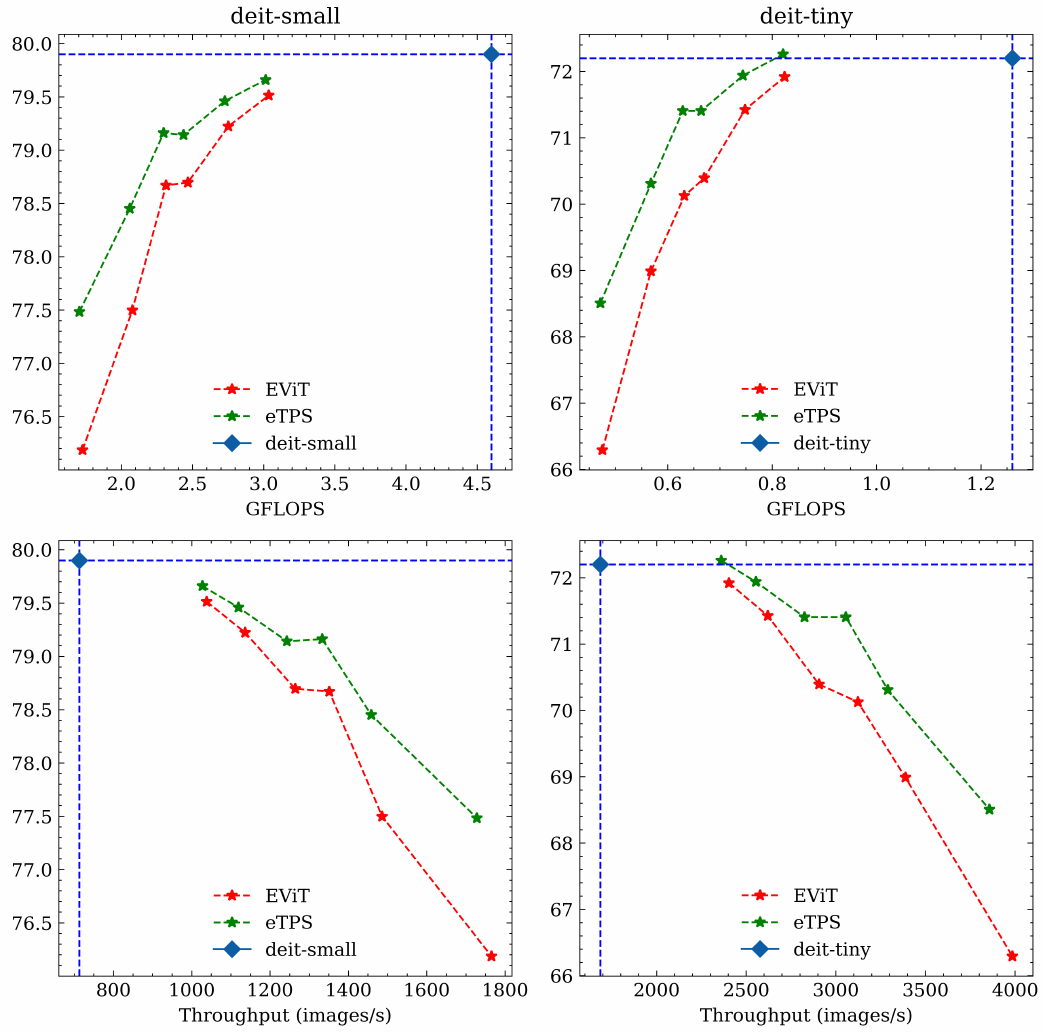}
        \caption{Comparison between our eTPS and EViT on DeiT.}
        \label{fig:dyn30-b}
    \end{subfigure}
    \caption{ImageNet1K results of our two variants: dTPS \& eTPS, and two baselines: DynamicViT \cite{rao2021dynamicvit} \& EViT \cite{liang2022not}, under different GFLOPs of pruned DeiT-small\&tiny\cite{touvron2021training}. The parameter number of two variants is the same as the two baselines respectively. The throughput is measured on a single NVIDIA RTX 2080Ti with a batch size of 32. The more aggressively we apply token pruning on backbones, the more competitive accuracy-computation trade-off our method shows. See supplementary materials for TPS on DeiT-base and with a $384\times 384$ input size .}
    \label{fig:dyn30}
\end{figure*}



\begin{figure*}
    \centering
    \includegraphics[width=0.75\linewidth]{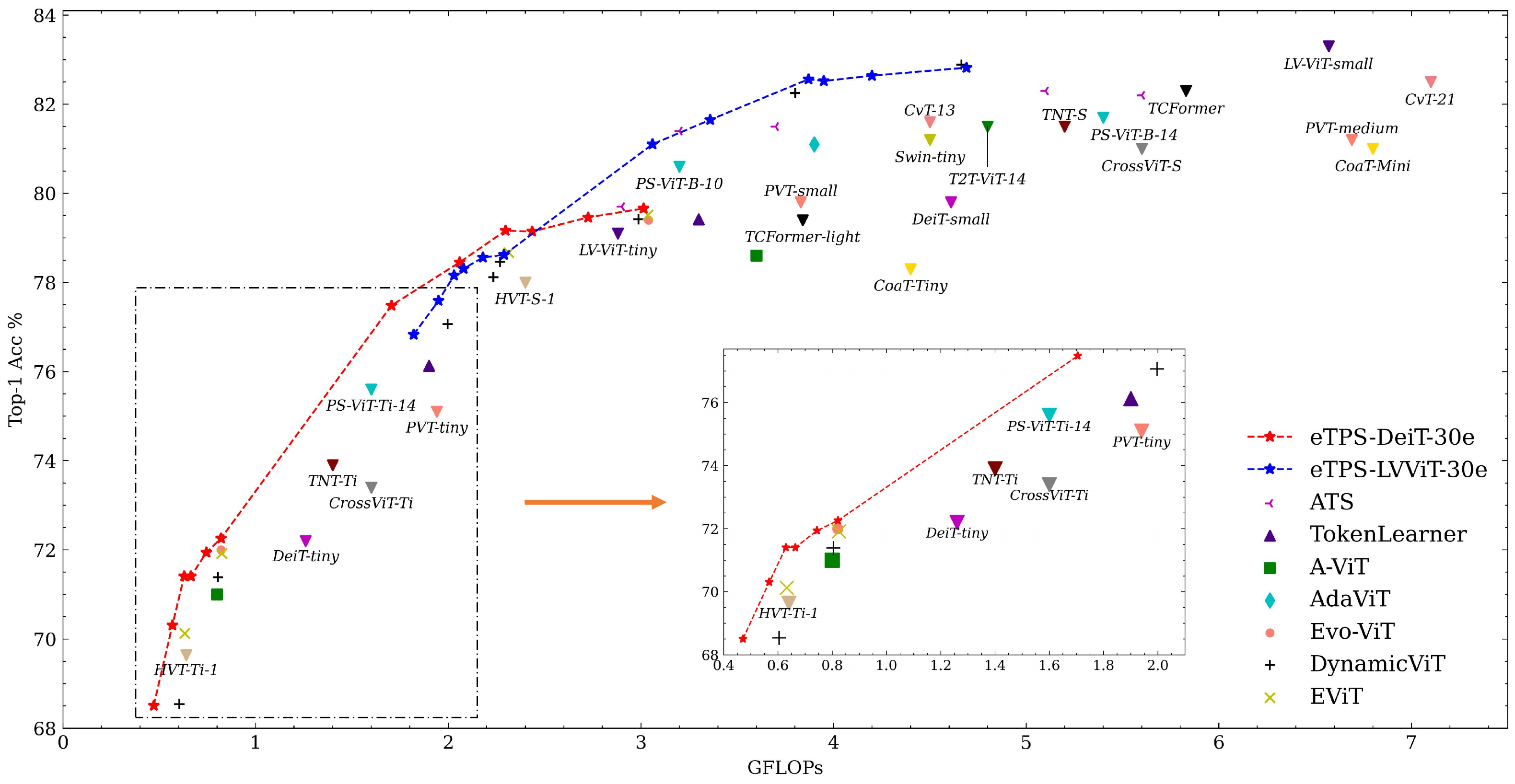}
    \caption{Comparison between DeiT \& LV-ViT with our TPS applied and other transformer methods, including token pruning~\cite{meng2022adavit,fayyazadaptive,rao2021dynamicvit,liang2022not,xu2022evo,meng2022adavit,han2021transformer}, vanilla ViTs~\cite{touvron2021training,jiang2021all,yue2021vision,yuan2021tokens,chen2021crossvit,pan2021scalable} and hybrid ViTs~\cite{liu2021swin,wang2021pyramid,zeng2022not,xu2021co}. Our TPS outperforms numerous state-of-the-art transformers on ImageNet1K image classification with only 30 fine-tuning epochs required.}
    \label{fig:all_exps}
\end{figure*}

\textbf{Datasets and evaluation metrics.} We conduct contrast experiments with two typical baselines: DynamicViT~\cite{rao2021dynamicvit} and EViT~\cite{liang2022not}, and compare our performances with state-of-the-art transformers. For quantitative comparisons, we report the Top-1 accuracy, the number of giga floating-point operations (GFLOPs), and throughput. The input size is set to $224 \times 224$ for all the experiments. The evaluated datasets include the ImageNet1K~\cite{deng2009imagenet} and the large fine-grained image classification dataset: iNaturalist 2019~\cite{fgvc6}. 

\textbf{Experiments Details.} We follow the same data augmentations used in DeiT~\cite{touvron2021training}. The model is initialized with pre-trained models' weights and  fine-tuned with different token pruning locations and keeping ratios. We adopt the AdamW~\cite{loshchilov2017decoupled} as the optimizer and a cosine learning rate scheduler. We compare our dTPS and eTPS with DynamicViT~\cite{rao2021dynamicvit} and EViT~\cite{liang2022not} under multiple pruning settings\footnote{The pruning settings include combinations of three multi-layer pruning settings: pruning locations include \{\nth{4},\nth{7},\nth{10}\},\{\nth{3},\nth{5},\nth{7},\nth{9}\}, and \{\nth{4},\nth{6},\nth{8},\nth{10}\}, and token keeping ratios $ \rho \in \{0.5,0.7\}$ }. We follow the same training settings and loss functions from \cite{liang2022not,rao2021dynamicvit}, except for the basic learning rate is set to $\frac{batch size}{1024}\times 2.5\times10^{-4}$ and no stage of fixing backbone weights in dynamicViT \& dTPS. The setting changes slightly because the training under the original setting appears unstable, especially with aggressive pruning.


\begin{figure}[t]
  \centering
  \includegraphics[width=1.0\linewidth]{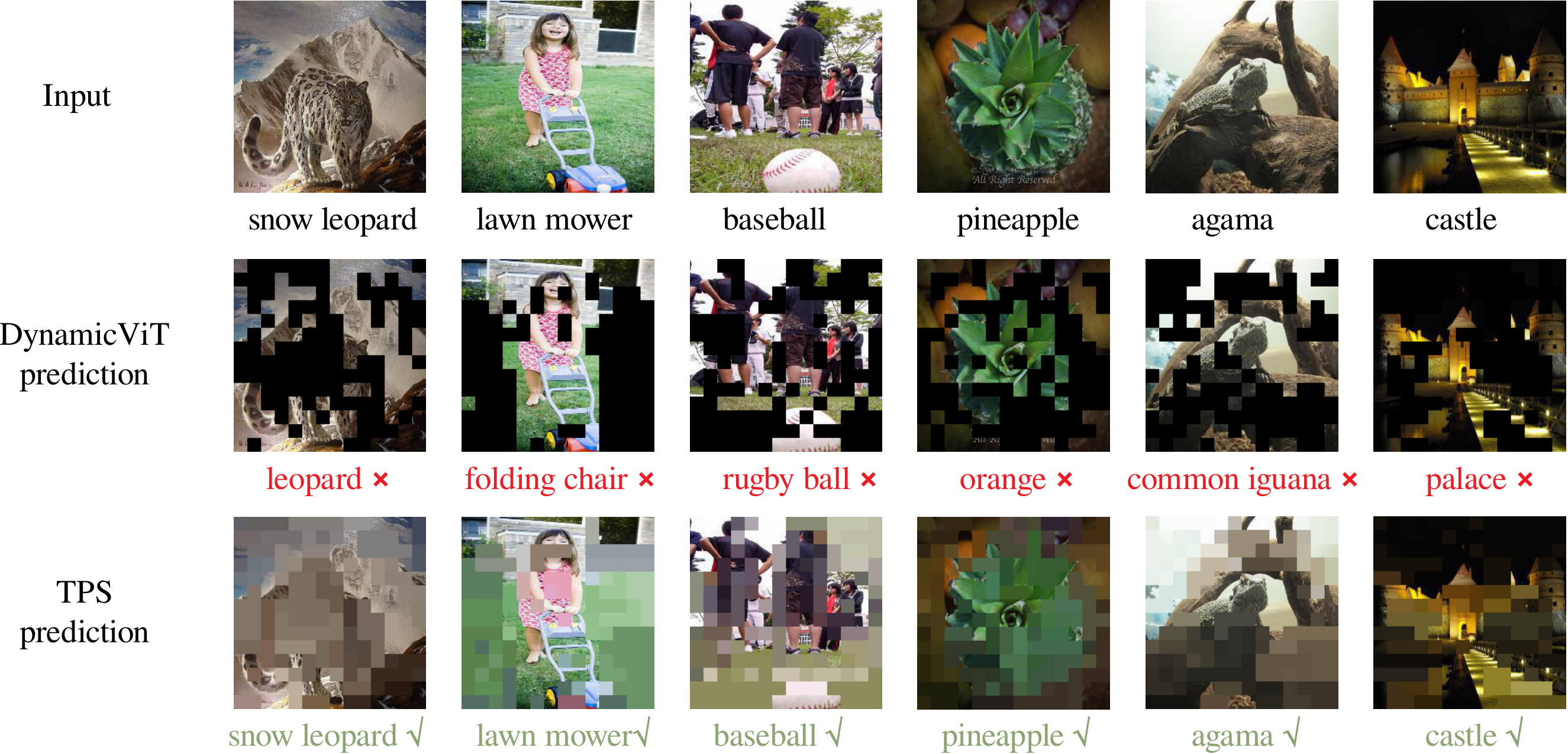}
  \caption{Visual comparisons between token pruning \cite{rao2021dynamicvit} and our TPS. DynamicViT and our dTPS on DeiT-small give the results. The pruning locations include \{\nth{4},\nth{7},\nth{8}\} and each pruning stage's keeping ratio is 0.5. The cases are displayed based on the results of the first pruning stage. For DynamicViT, the blank areas denote pruned tokens; for dTPS, we mask each group of matched tokens as the same color for visualization clarity.}
  \label{fig:cmp_cases}
\end{figure}

\subsection{Main Results}

\textbf{Comparison to baselines.} As ~\cref{fig:dyn30} shows, we compare our method with the token pruning baseline: DynamicViT, and the token reorganization baseline: EViT, by replacing their original pruning modules with our dTPS \& eTPS modules. The contrast experiments involve DeiT-small\&tiny. All the models in this part are fine-tuned 30 epochs under multiple pruning settings. Under all the settings, our method outperforms DynamicViT and EViT. Both dynamicViT and EViT encounter a larger accuracy drop along with the progressively aggressive pruning. While shrinking the computational budgets of DeiT to 35\%, our method can avoid \textbf{1\%-6\%} accuracy decline compared with baselines. Equipped with our TPS, we can accelerate the throughput of DeiT-small to \textbf{1745 images/s}, which is beyond that of DeiT-tiny: \textbf{1686 images/s}, and surpass the accuracy of DeiT-tiny by \textbf{4.78\%}.

\textbf{Visual Comparisons.} We demonstrate the cases from ImageNet1K~\cite{deng2009imagenet}, which DeiT predicts correctly at first but gives the wrong predictions after being applied with token pruning. As \cref{fig:cmp_cases} shows, the imperfect pruning policy brings the context information loss, which leads to a close but incorrect prediction. However, our TPS remedies these cases by saving the pruned tokens' information. 

\textbf{Comparison to states of the art.} In \cref{fig:all_exps}, we demonstrate our TPS performances compared with other state-of-the-art transformers, including token pruning methods~\cite{meng2022adavit,fayyazadaptive,rao2021dynamicvit,liang2022not,xu2022evo,meng2022adavit}, vanilla ViTs~\cite{touvron2021training,jiang2021all,yue2021vision,yuan2021tokens,chen2021crossvit,pan2021scalable} and hybrid ViTs~\cite{liu2021swin,wang2021pyramid,zeng2022not,xu2021co}. By integrating DeiT-small\&tiny and LV-ViT-small\&tiny with TPS and fine-tuning them only 30 epochs, we can achieve a quite competitive performance among numerous vision transformers from the perspective of accuracy-computation trade-off.

\textbf{Extension on more backbones.}
As shown in \cref{tab:vanilla_vits } and \cref{tab:hybrid_vits}, we incorporate TPS into different vanilla ViTs~\cite{touvron2021training,yue2021vision,yuan2021tokens} and hybrid ViTs~\cite{wang2021pyramid,zeng2022not} to prove the flexibility and generalization. For vanilla ViTs, our TPS outperforms EViT~\cite{liang2022not}, Evo-ViT~\cite{xu2022evo}, A-ViT~\cite{yin2022vit}, IA-RED$^2$ and SPViT~\cite{kong2022spvit} with equal or slightly increasing computation while using DeiT~\cite{touvron2021training}, LV-ViT~\cite{jiang2021all} as backbones. DeiT-small\&tiny with TPS applied can surpass the pre-trained models by \textbf{0.3\%} and \textbf{0.7\%} in accuracy under 100 fine-tuning epochs. For hybrid ViTs, we can compress the GFLOPS of PVT-tiny by 13\% and improve its accuracy by 0.1\%. 

\begin{table}
\centering
\scalebox{0.8}{
\begin{tabular}{lccc}\toprule[2pt]
Method &Param(M) &GFLOPs &Top-1 Acc.(\%) \\\midrule
DeiT-S\ &22.05 &4.6 &79.8 \\\midrule
DynamicViT\cite{rao2021dynamicvit} &22.77 &2.9 &79.3 \\
EViT\cite{liang2022not} &\textbf{22.05} &3.0 &79.5 \\
ATS$^\dag$\cite{fayyazadaptive} &\textbf{22.05} &2.9 &\textbf{79.7} \\
A-ViT$^\dag$\cite{yin2022vit} (100 epochs) &\textbf{22.05} &3.6 &78.6 \\
Evo-ViT\cite{xu2022evo} (300 epochs) &\textbf{22.05} &3.0 &79.4 \\
SPViT\cite{kong2022spvit} (75 epochs) &22.13 &\textbf{2.7} &79.3 \\
IA-RED$^2$\cite{pan2021ia} (90 epochs) &- &- &79.1 \\
\cellcolor[HTML]{f3f3f3}eTPS (ours) &\cellcolor[HTML]{f3f3f3}\textbf{22.05} &\cellcolor[HTML]{f3f3f3}3.0 &\cellcolor[HTML]{f3f3f3}\textbf{79.7} \\
\cellcolor[HTML]{f3f3f3}dTPS* (ours) &\cellcolor[HTML]{f3f3f3}22.77 &\cellcolor[HTML]{f3f3f3}3.0 &\cellcolor[HTML]{f3f3f3}\textbf{80.1} \\\midrule[2pt]
DeiT-T &\textbf{5.72} &1.3 &72.2 \\\midrule
DynamicViT(re-impl)\cite{rao2021dynamicvit} &5.90 &\textbf{0.8} &71.4 \\
EViT(re-impl)\cite{liang2022not} &5.72 &\textbf{0.8} &71.9 \\
A-ViT$^\dag$\cite{yin2022vit} (100 epochs) &\textbf{5.00} &\textbf{0.8} &71.0 \\
Evo-ViT\cite{xu2022evo} (300 epochs) &5.72 &\textbf{0.8} &72.0 \\
SPViT\cite{kong2022spvit} (75 epochs) &- &0.9 &72.1 \\
\cellcolor[HTML]{f3f3f3}eTPS (ours) &\cellcolor[HTML]{f3f3f3}5.72 &\cellcolor[HTML]{f3f3f3}\textbf{0.8} &\cellcolor[HTML]{f3f3f3}\textbf{72.3} \\
\cellcolor[HTML]{f3f3f3}dTPS* (ours) &\cellcolor[HTML]{f3f3f3}5.90 &\cellcolor[HTML]{f3f3f3}\textbf{0.8} &\cellcolor[HTML]{f3f3f3}\textbf{72.9} \\\midrule[2pt]
LV-ViT-S &26.17 &6.6 &83.3 \\\midrule
DynamicViT\cite{rao2021dynamicvit} &26.89 &\textbf{3.8} &82.0 \\
EViT\cite{liang2022not} &\textbf{26.17} &3.9 &\textbf{82.5} \\
\cellcolor[HTML]{f3f3f3}eTPS (ours) &\cellcolor[HTML]{f3f3f3}\textbf{26.17} &\cellcolor[HTML]{f3f3f3}\textbf{3.8} &\cellcolor[HTML]{f3f3f3}\textbf{82.5} \\
\cellcolor[HTML]{f3f3f3}dTPS* (ours) &\cellcolor[HTML]{f3f3f3}26.89 &\cellcolor[HTML]{f3f3f3}\textbf{3.8} &\cellcolor[HTML]{f3f3f3}\textbf{82.6} \\\midrule[2pt]
LV-ViT-T &8.53 &2.9 &79.1 \\\midrule
DynamicViT(re-impl)\cite{rao2021dynamicvit} &8.82 &2.0 &77.1 \\
\cellcolor[HTML]{f3f3f3}eTPS (ours) &\cellcolor[HTML]{f3f3f3}\textbf{8.53} &\cellcolor[HTML]{f3f3f3}\textbf{2.0} &\cellcolor[HTML]{f3f3f3}\textbf{78.0} \\
\cellcolor[HTML]{f3f3f3}dTPS* (ours) &\cellcolor[HTML]{f3f3f3}\textbf{8.82} &\cellcolor[HTML]{f3f3f3}\textbf{2.0} &\cellcolor[HTML]{f3f3f3}\textbf{78.7} \\\midrule[2pt]
PS-ViT-B/14\cite{yue2021vision} &21.34 &5.4 &81.7 \\\midrule
ATS$^\dag$\cite{fayyazadaptive} &\textbf{21.34} &\textbf{3.7} &\textbf{81.5} \\
\cellcolor[HTML]{f3f3f3}dTPS* (ours) &\cellcolor[HTML]{f3f3f3}22.07 &\cellcolor[HTML]{f3f3f3}\textbf{3.7} &\cellcolor[HTML]{f3f3f3}\textbf{81.5} \\
\bottomrule[2pt]
\end{tabular}
}
\caption{Comparison among different token pruning methods applied to multiple vanilla vision transformers. ``*" denotes our method is fine-tuned 100 epochs. Methods marked with "\dag" do not support constant-shape inference. Prior methods above are trained 30 epochs by default unless otherwise specified. ``Re-impl" means that we implement the method according to the official code. For a fair comparison with prior methods, we utilize computationally comparable pruning setups to fine-tune backbones with TPS.}\label{tab:vanilla_vits }
\end{table}
\begin{table}[!htp]
\centering
\scalebox{0.8}{\begin{tabular}{lccc}\toprule
Method &Param (M) &GFLOPs &Top-1 Acc. (\%) \\\midrule
PVT-T\cite{wang2021pyramid} &13.23 &1.94 &75.1 \\
dTPS* (ours) &13.85 &\textbf{1.69 (-13\%)} &\textbf{75.2 (+0.1)} \\\midrule
PVT-S &24.49 &3.83 &79.8 \\
dTPS* (ours) &25.11 &\textbf{3.14 (-18\%)} &79.2 (-0.6) \\\midrule
CvT-13\cite{wu2021cvt} &20.00 &4.58 &81.6 \\
dTPS* (ours) &20.72 &\textbf{3.04 (-34\%)} &80.8 (-0.8) \\\midrule
CvT-21 &31.62 &7.21 &82.5 \\
dTPS* (ours) &32.35 &\textbf{4.10 (-43\%)} &80.9 (-1.6) \\
\bottomrule
\end{tabular}}
\caption{Experiments of our methods applied to hybrid vision transformers, including PVT~\cite{wang2021pyramid} and CVT~\cite{wu2021cvt}.}\label{tab:hybrid_vits}

\end{table}

\textbf{Fine-Grained Visual categorization.} We compare our dTPS with DynamicViT by fine-tuning DeiT on iNaturalist 2019~\cite{fgvc6} as shown in Tab.~\ref{inat_result}. See the supplementary materials for the training details on iNaturalist 2019~\cite{fgvc6}. Compared with dynamicViT, our dTPS obtains 0.3\% accuracy improvement in DeiT-tiny and 0.2\% accuracy improvement in DeiT-small when fine-tuning 30 epochs, respectively. We further fine-tune dTPS 100 epochs and observe a significant improvement in both backbones. Notably, dTPS fine-tuned with 100 epochs is only 0.1\% lower than Deit-small while shrinking the computational budgets of Deit-small to 65\%. 

\begin{table}
  \centering
  \scalebox{0.8}{
  \begin{tabular}{lccc}
    \toprule[2pt]
    Method & Param(M) & GFLOPs & Top-1 Acc.(\%) \\
    \midrule[2pt]
    DeiT-S\cite{touvron2021training}  &  22.05 & 4.6 & 74.8  \\
    \hline
    DynamicViT(re-impl)\cite{rao2021dynamicvit} & \textbf{22.77} & \textbf{2.9} & 74.0 \\
    \cellcolor[HTML]{f3f3f3}dTPS (ours) &\cellcolor[HTML]{f3f3f3}\textbf{22.77} &\cellcolor[HTML]{f3f3f3}3.0 &\cellcolor[HTML]{f3f3f3}74.2 \\
    \cellcolor[HTML]{f3f3f3}dTPS* (ours) &\cellcolor[HTML]{f3f3f3}\textbf{22.77} &\cellcolor[HTML]{f3f3f3}3.0 &\cellcolor[HTML]{f3f3f3}\textbf{74.7} \\
    \midrule[2pt]
    DeiT-T &  5.72 & 1.26 & 72.8  \\
    \hline
    DynamicViT(re-impl)\cite{rao2021dynamicvit} & \textbf{5.90} & \textbf{0.8} & 71.4 \\
    \cellcolor[HTML]{f3f3f3}dTPS (ours) &\cellcolor[HTML]{f3f3f3}\textbf{5.90} &\cellcolor[HTML]{f3f3f3}\textbf{0.8} &\cellcolor[HTML]{f3f3f3}71.7 \\
    \cellcolor[HTML]{f3f3f3}dTPS* (ours) &\cellcolor[HTML]{f3f3f3}\textbf{5.90} &\cellcolor[HTML]{f3f3f3}\textbf{0.8} &\cellcolor[HTML]{f3f3f3}\textbf{72.4} \\
    \bottomrule[2pt]
  \end{tabular}}
  \caption{Results of dynamicViT\cite{rao2021dynamicvit} and dTPS on iNaturalist 2019~\cite{fgvc6}. The two models are trained 30 epochs by default. ``*" denotes the model is trained 100 epochs. ``Re-impl" means we implement the method on the backbone according to the official code. See the appendix for the training details of the backbone.}
  \label{inat_result}
\end{table}

\subsection{Ablation Study}
\label{ablation_study}

\begin{figure}[t]
    \centering
    \includegraphics[width=0.8\linewidth]{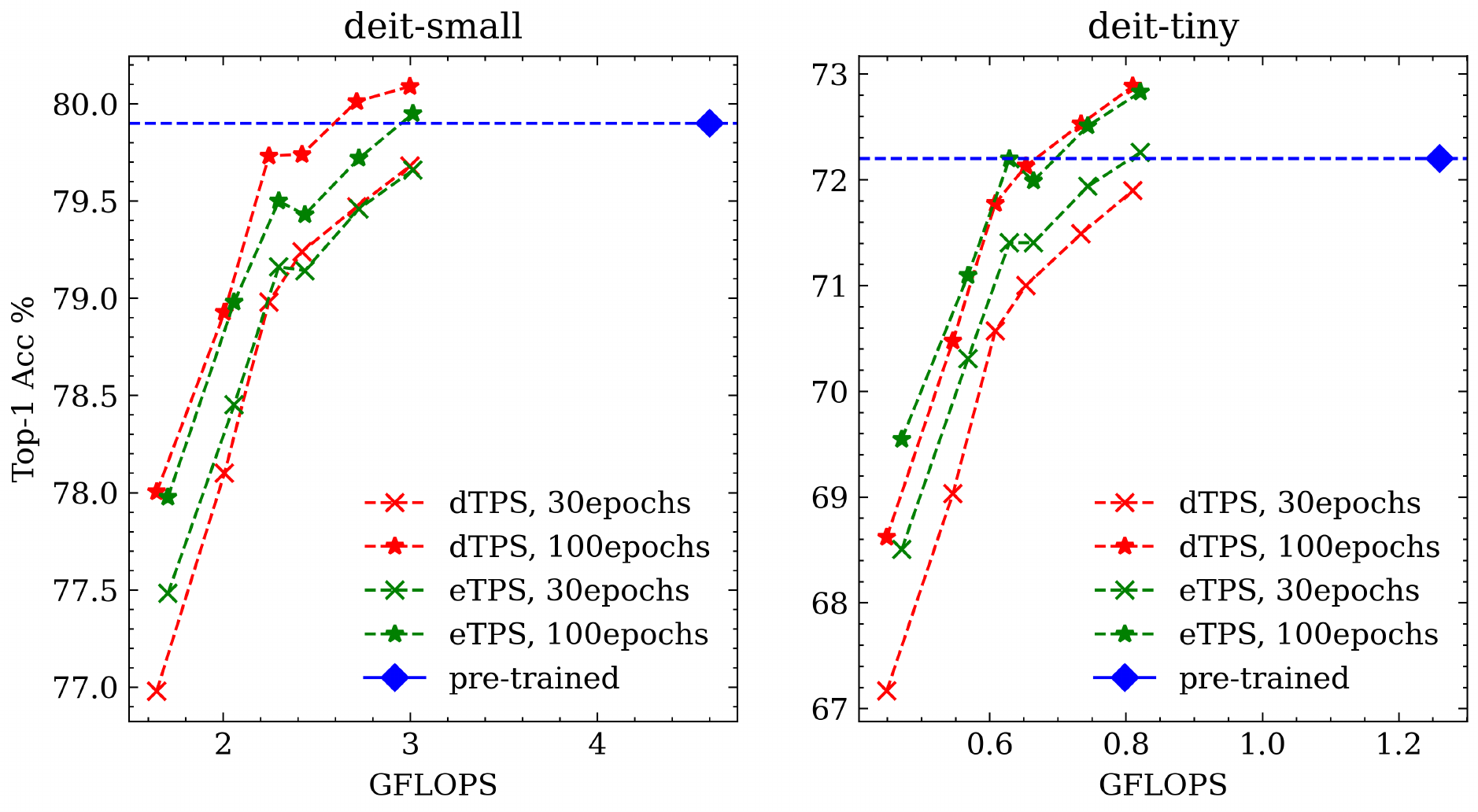}
    \caption{ImageNet1K results of our dTPS \& eTPS under different GFLOPs of the pruned model under 30 \& 100 fine-tuing epochs. Both dTPS and eTPS benefit from longer epochs and surpass the performances of pre-trained DeiT-small\&tiny. }
    \label{fig:100e}
\end{figure}



\begin{table}[t]

     \centering
    \scalebox{0.8}{
     \begin{tabular}{lcc}\toprule
    Feature Type &Top-1 Acc. (\%) \\\midrule
    Full &\textbf{71.90} \\
    Content &71.73 \\
    Position &70.92 \\
    \bottomrule
    \end{tabular}
    }
    \caption{Different feature types used in the matching step. Conducted on DeiT-tiny with pruned layers at \{\nth{4},\nth{7},\nth{8}\} and keeping ratio set to 0.7.}\label{feat_exps}

\end{table}



\begin{table}[t]

    \centering
    \scalebox{0.8}{
    \begin{tabular}{lcccc}\toprule
    TPM Variant & Similarity Matrix & GFLOPs & Top-1 Acc.(\%) \\\midrule
    \multirow{2}{*}{dTPS} &Cosine similarity &0.810 &\textbf{71.90} \\
    &Previous attention &0.807 &71.35 \\
    \multirow{2}{*}{eTPS} &Cosine similarity &0.821  &\textbf{72.26} \\
    &Previous attention &0.818 &71.67 \\
    \bottomrule
    \end{tabular}}
    \caption{Comparison between different types of cost matrix. The baseline denotes calculating the tokens' cosine similarity; the previous attention denotes that we reuse it to devise the matching results. Conducted on DeiT-tiny with pruned layers at \{\nth{4},\nth{7},\nth{8}\} and keeping ratio set to 0.7.}\label{reuse_attn}

\end{table}

\textbf{Epochs of training.} Fig.~\ref{fig:100e} shows that both variants can benefit from longer training epochs and surpass the DeiT-small\&tiny with only 65\% GFLOPS. However, the benefit of epochs varies slightly in two variants. Because the class-token attention scoring requires no extra optimization target, eTPS performs better than dTPS under 30 epochs. On the other hand, dTPS can benefit more from longer training epochs in DeiT-small, for its learnable scoring brings higher performance upper bound.


\textbf{Feature Type.} We show the effects of feature type used to establish the matching relationships. Supposing $\boldsymbol{x}_i$ is the full embedding of the token and the position feature $\boldsymbol{p}_i$ is the corresponding positional embedding, we define the content feature as $\boldsymbol{x}_i-\boldsymbol{p}_i$. As Tab.~\ref{feat_exps} illustrates, the entire feature is more favorable for it contains both the content and position information.

\textbf{Similarity Matrix.} Considering that the dot-product attention of query and key measures tokens' relationships naturally, we have tried reusing the previous attention to replace the computations of cosine similarities in the matching step. We believe the previous attention is outdated to measure current tokens' relations and \cref{reuse_attn} shows that calculating the cosine similarity of current features outperforms reusing the attention with only a minor computational increase. 


\subsection{Robustness Experiments}


We generate random token selection policies to construct manufactured policy errors that simulate the cases brought by sub-optimal token pruning strategies. All models are based on DeiT-small and fine-tuned 30 epochs with identical pruning setups. We run the experiments under random policies 100 times and report the average results. By comparing the performances of our method with dynamicViT~\cite{rao2021dynamicvit} and EViT~\cite{liang2022not}, the accuracy drop from the original to random policies denotes the robustness under incorrect policies. As shown in \cref{random_policy}, our inter-block version dTPS and intra-block version eTPS have fewer accuracy drops than dynamicViT \cite{rao2021dynamicvit} and EViT \cite{liang2022not}.

\begin{table}
    \centering
    
    \scalebox{0.8}{
    \begin{tabular}{lccc}\toprule
    Methods & Policy & Top-1 Acc. ($\%$)\\
    \hline
    \multirow{2}{*}{DynamicViT} &Original &79.42 \\
    &Random &76.51 (-3.7) \\
    \multirow{2}{*}{dTPS} &Original &79.68 \\
    &Random &78.19 (\textbf{-1.9}) \\
    \hline
    \multirow{2}{*}{EViT} &Original &79.51 \\
    &Random &77.47 (-2.6) \\
    \multirow{2}{*}{eTPS} &Original &79.66 \\
    &Random &78.06 (\textbf{-2.0}) \\
    \bottomrule
    \end{tabular}
    }
    
    \caption{ImageNet1K results of applying random token selection policy to our methods and baselines. The percentages in parentheses represent the relative performance degradation ratio brought by random policies.}
  \label{random_policy}
\end{table}


\section{Conclusions and Limitations}
In this paper, we presented a novel joint Token Pruning \& Squeezing (TPS) module to compress vision transformers more aggressively. With the capability of conserving information, our TPS can avoid a significant performance drop compared to token pruning and reorganization. Our method has better efficiency than prior token pruning methods and states of the arts in vision transformers. Extensive experiments under various backbones and quantitative analyses show our flexibility and robustness. 

However, there are still some limitations to our method. Firstly, structured spatial operations of hybrid ViTs restrict the straightforward integration of token pruning. Secondly, the procedure of fine-tuning pre-trained models might be replaced by more advanced pruning-aware training-from-scratch schemes to shorten the total training time. In the future, we will evolve our method to be more adaptive to hybrid ViTs and apply it to more dense prediction tasks.


\setcounter{section}{0}
\leftline{\textbf{\large{Supplementary Material}}}

\begin{figure*}
    \centering
    \begin{subfigure}{0.4\linewidth}
        \includegraphics[width=0.95\linewidth]{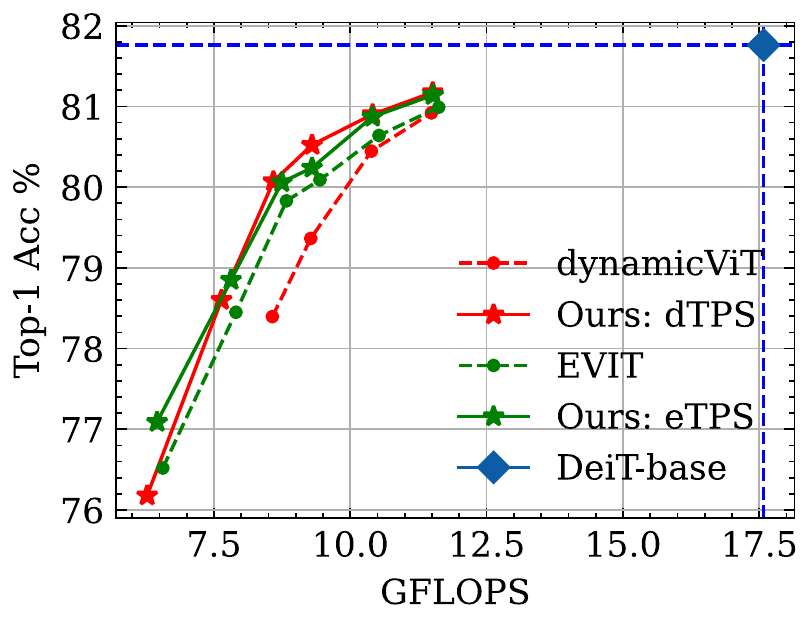}
        \caption{Comparison on DeiT-B.}
        \label{DeiT_base}
    \end{subfigure}
    \hspace{7mm}
    \begin{subfigure}{0.4\linewidth}
        \includegraphics[width=1.0\linewidth]{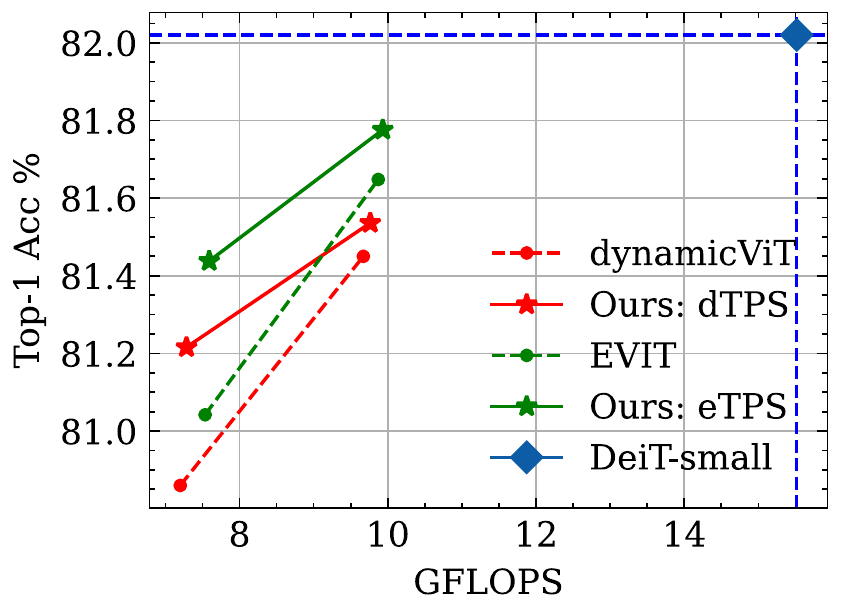}
        \caption{Comparison on DeiT-S-384$\times$384.}
        \label{DeiT_384}
    \end{subfigure}
    \caption{ Note that dynamicViT can't converge in the two most aggressive pruning settings in (b).}
    \label{fig:larger}
\end{figure*}

\begin{figure*}[htbp]
  \centering
  \includegraphics[width=0.8\linewidth]{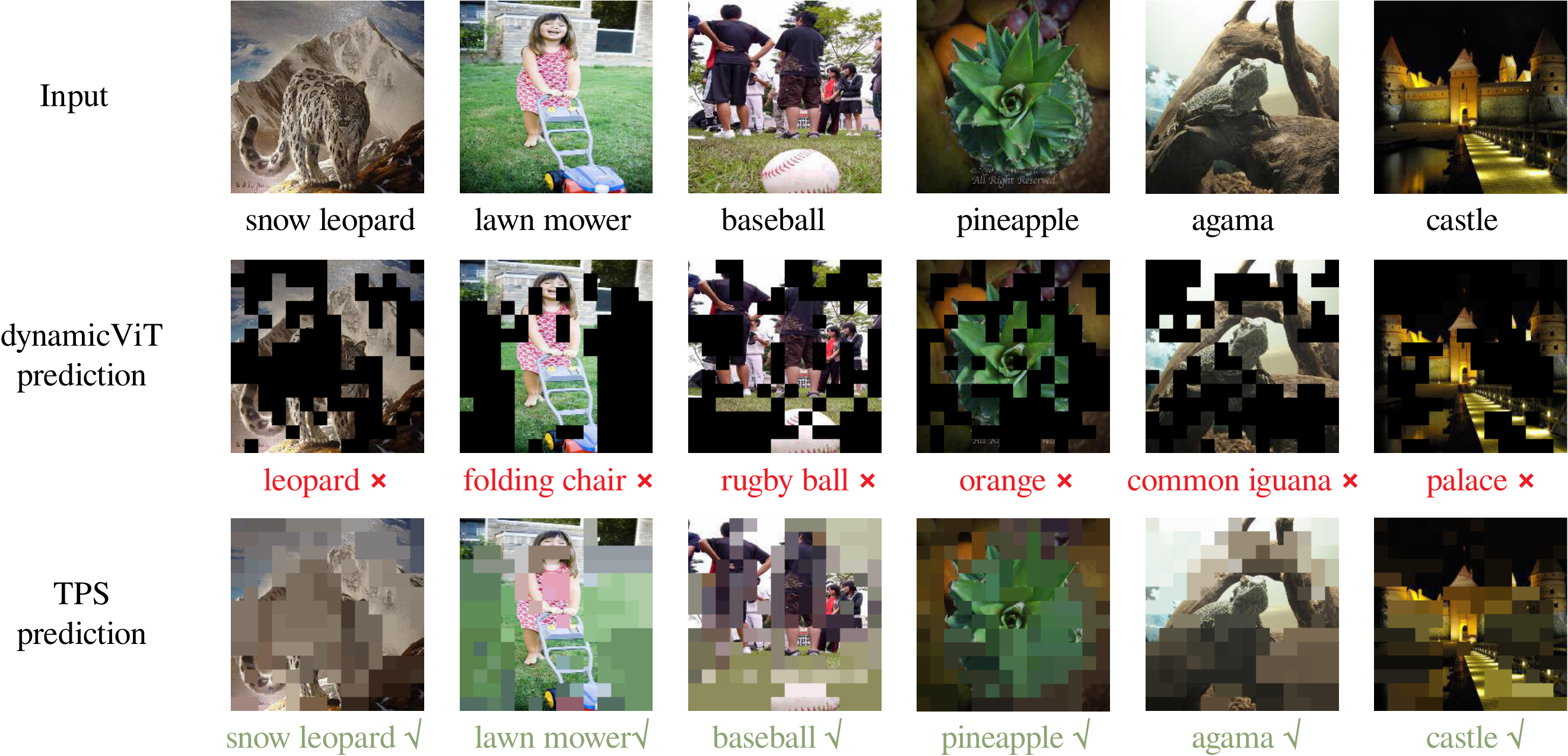}
  
  \includegraphics[width=0.8\linewidth]{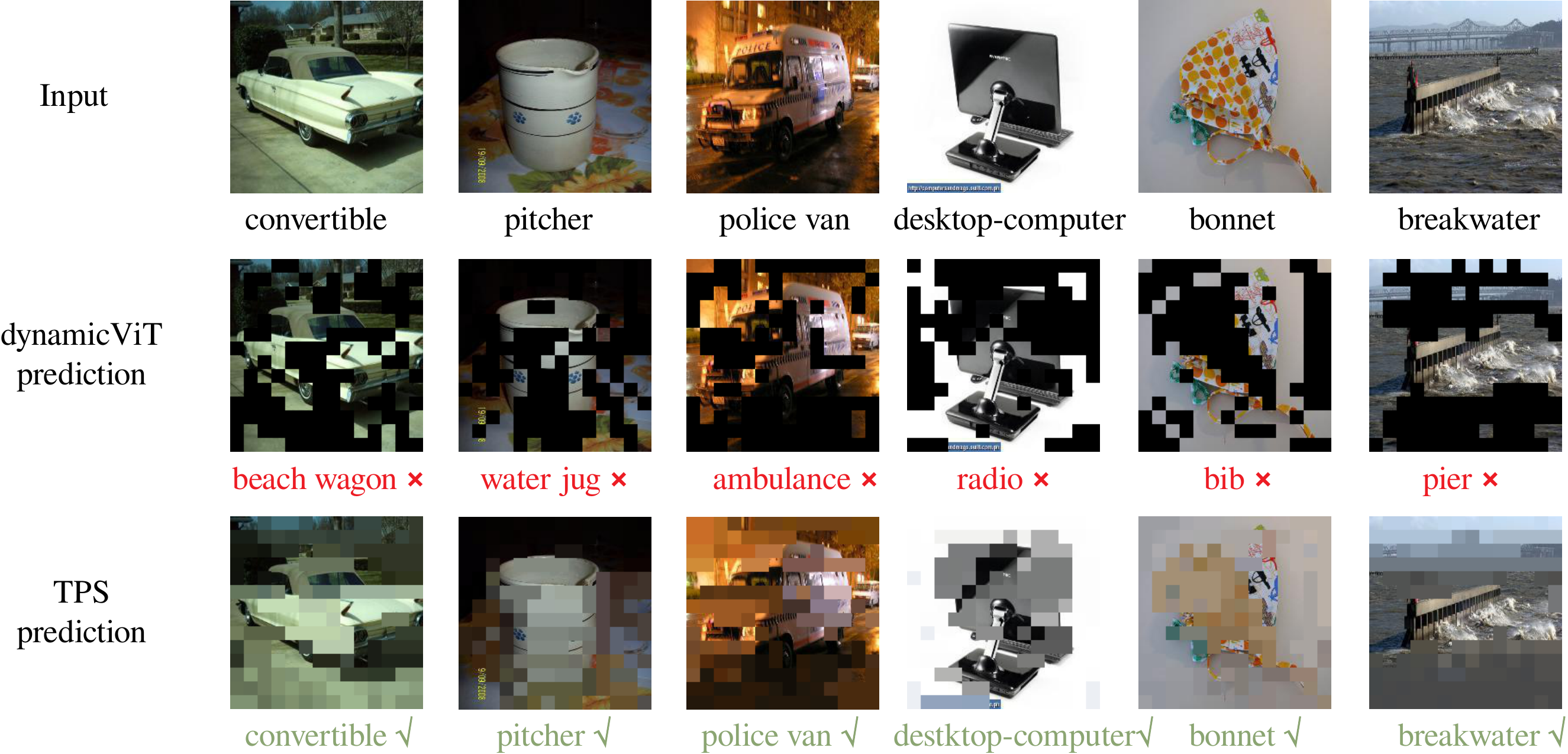}
  \caption{Comparisons between dynamicViT ~\cite{rao2021dynamicvit} and our joint Token Pruning \& Squeezing. The results are given by dynamicViT-DeiT-small and our TPS-DeiT-small with the same pruning setup ( pruning locations are \{\nth{4},\nth{7},\nth{8}\} and token keeping ratio of each pruning stage $\rho$ is 0.5 ). The cases of dynamicViT and our TPS are displayed based on practical operations of the first stage with pruning applied. For dynamicViT, the blank tokens denote pruned ones; for our TPS, we mask each group of matched tokens in the TPS as the same color for visualization clarity. }
  \label{fig:showcases}
\end{figure*}

\begin{table*}
    \centering
    \begin{subtable}[t]{0.45\linewidth}
    \centering
        \scalebox{1.0}{
            \begin{tabular}{lcc}\toprule
            Matching Method &Acc. (\%) \\\midrule
            N:1 &\textbf{71.90} \\
            1:1  &69.02 \\
            \bottomrule
            \end{tabular}
        }
        \caption{Different matching methods on dTPS-DeiT-T. The keep ratio is \textbf{0.7}.}
        \label{matching}
    \end{subtable}
    \hspace{5mm}
    \begin{subtable}[t]{0.45\linewidth}
        \centering
        \scalebox{1.0}{
            \begin{tabular}{lccc}\toprule
                Fusing Method & Policy & Acc. ($\%$)\\
                \hline
                \multirow{2}{*}{Weighting} &Original &70.58 \\
                &Random & \textbf{65.56 (-5.02)} \\
                \multirow{2}{*}{Average} &Original &70.47 \\
                &Random &65.173 (-5.30) \\
                \bottomrule
                \end{tabular}
        }
        \caption{Robustness comparison of fusing methods on dTPS-DeiT-T. The keep ratio is \textbf{0.5}. }
        \label{fusing}
    \end{subtable}
    \caption{ The pruned layers includes \nth{4},\nth{7},\nth{8}. (a) N:1 matching employed by TPS finds the nearest reserved token for each pruned token to inject multiple tokens into the same token, while 1:1 matching finds the nearest pruned token for each reserved token. (b) The similarity-weighting fusing obtains a lower accuracy drop under random token squeezing (see more details in the main paper: Sec.~4.3 ) than average fusing. } \label{ablation}
\end{table*}

\section{Overview}

In the supplemental materials, we show the following details of our joint Token Pruning \& Squeezing (TPS):
\begin{itemize}
\item Visualizations.
    \item Details of two variants.
    \item TPS on hybrid ViTs.
    \item Detailed experiment settings.
    \item TPS on larger models and with larger input size.
    \item TPS under different keep ratios.
    \item More ablations about TPS design. 
    
\end{itemize}

\section{Visualizations}
We demonstrate the additional cases from ImageNet1K~\cite{deng2009imagenet}, which our TPS-DeiT and DeiT predict correctly but dynamicViT-DeiT predict wrongly. As Fig~\ref{fig:showcases} shows, we found that the imperfect pruning policy brings the loss of background context and incomplete subject, which puzzles the model and leads to a close but incorrect prediction. However, our TPS conquers these cases by squeezing the information of pruned tokens into similar reserved tokens. 

\section{Details of Two Variants}
We design two variants of TPS: dTPS and eTPS, to show our flexibility and compare fairly with dynamicViT~\cite{rao2021dynamicvit}
and EViT~\cite{liang2022not}. Theoretically, our TPS can be incorporated with any token pruning method. In this paper, we choose dynamicViT and EViT as baselines for their strong performance and concise forms. The major disparities between the two variants are as follows:

\textbf{Forward procedure.} The TPS module drops the tokens from inputs practically except for the training stage of dTPS. In each pruning stage, the dTPS module employs the gumbel-softmax~\cite{jang2016categorical} to sample binary decision mask randomly during training and maintain the presently reserved mask and pruned mask to avoid previously pruned tokens from participating in the matching and fusing step. In the subsequent attention layer, the attention masking strategy from dynamicViT~\cite{rao2021dynamicvit} is employed to erase the effects of dropped tokens. The implementation details can be found in the code file.

\textbf{Position to insert.} As mentioned in the paper, dTPS and eTPS cut down tokens in inter-block and intra-block ways, respectively. The dTPS module is inserted before the transformer block, while the eTPS module is inserted after the multi-head attention layer. The distinction derives from the different token scoring methods. The learnable score prediction employed by dynamicViT and dTPS does not rely on any internal operation of transformer blocks, while the scoring based on the class-token attentions requires the results from the multi-head attention block.

\textbf{Parameters.} The eTPS module is parameter-free, while the dTPS module increases the total number of parameters by a small amount due to its learnable token score prediction head.

\textbf{Performances.} The performances of dTPS and eTPS modules are close but can be slightly different when training epochs changes. According to our experiments, the eTPS module outperforms the dTPS module under 30 epochs. The opposite results were observed under 100 epochs. The difference demonstrates that extra parameters of dTPS modules endow the model with a higher upper limit.

\section{TPS on Hybrid ViTs}
We conduct experiments on PVT~\cite{wang2021pyramid} and CvT~\cite{wu2021cvt} to prove our design is compatible with hybrid ViTs.

\subsection{PVT}
    Generally, We insert dTPS modules between the patch embedding layer and the subsequent basic block of each pruned stage. Unlike TPS on vanilla ViTs, we reserve attentive tokens from the whole tokens set in each pruned stage and utilize the masking or padding operations to maintain the complete spatial structure during training and inference, respectively.
    
    \textbf{Training.} The spatial reduction layer of the basic block in PVT requires input with a complete spatial structure. For the basic block with a spatial reduction block, given the policy $M$, we maintain the complete spatial structure and mask the dropped tokens in key $K$ and in value $V$ with zeros before the spatial reduction layer as follows:
    \begin{small}
    \begin{equation}
        S\!R\!A^*(Q,K,V)=M\!H\!A(Q,S\!R(K \odot M),S\!R(V \odot M))
    \end{equation}
    \end{small}
    Here, $S\!R\!A^*$ is the modified spatial reduction attention, $M\!H\!A$ is the multi-head attention operation, and $SR$ is the spatial reduction layer. Moreover, we perform the same masking operation on dropped tokens before the patch embedding layer of next stage. 
    
    For the basic block without a spatial reduction layer, no masking operation is needed ,and we conduct the attention masking strategy from dynamicViT~\cite{rao2021dynamicvit} to erase the effects of the dropped tokens.
    
    \textbf{Inference.} The inference procedure of dTPS is adjusted slightly to practically accelerate the spatial reduction layer. The input tokens are pruned by a top-k selection operation based on the scoring results. For the block with a spatial reduction layer, we pad the previously dropped tokens with zero in the $S\!R\!A^*$ to maintain the complete spatial structure. 
    \begin{small}
    \begin{equation}
        K'=Pad(K,M)
    \end{equation}
    \begin{equation}
        V'=Pad(V,M)
    \end{equation}
    \begin{equation}
        S\!R\!A^*(Q,K,V)=M\!H\!A(Q,S\!R(M'),S\!R(V'))
    \end{equation}
    \end{small}
    Also, the same padding operation is utilized before the patch embedding layer of the next stage. For the block without a spatial reduction layer, no padding operation is needed either. The requirement of complete spatial structure leads to less shrinkage of computations.


\subsection{CvT}
    The last stage of CvT~\cite{wu2021cvt} contains most of its blocks; therefore we only modify the last stage with our dTPS. The operations remain the same for other stages as the original CvT~\cite{wu2021cvt}.
    
    \textbf{Training.} The convolutional projection operation in CvT requires the input with a complete spatial structure. Given the policy $M$, we mask the dropped tokens with zeros before the convolution projection:
    \begin{equation}
        Q,K,V = Convolutional Projection(X \odot M)
    \end{equation}

    \textbf{Inference.} The input tokens are pruned by a top-k selection operation based on the scoring results. To maintain the complete spatial structure, we pad the previously dropped tokens with zeros in the convolutional projection layer.
    \begin{equation}
        X' = Pad(X, M)
    \end{equation}
    \begin{equation}
        Q,K,V = Convolutional Projection(X')
    \end{equation}
    
    ATS~\cite{fayyazadaptive} conducts experiments on CvT~\cite{wu2021cvt} as well. It takes a variant of CvT~\cite{wu2021cvt} as the pre-trained model without convolutional projection in stage 3. It only performs token pruning in stage 3 to avoid the extra operation to maintain the structured spatial input. Compared to ATS~\cite{fayyazadaptive}, our method utilizes masking and padding during training and inference to keep the spatial structure.


\section{Detailed Experiment Settings}
\subsection{ImageNet-1K Classification}
    All experiments follow the same data augmentations used in DeiT\footnote{ \url{https://github.com/facebookresearch/DeiT}}~\cite{touvron2021training}. All the model is initialized with pre-trained models' weights and  fine-tuned with different token pruning location and token keeping ratio. We adopt the AdamW~\cite{loshchilov2017decoupled} as the optimizer and a cosine learning rate scheduler. 
    
    \textbf{TPS on DeiT~\cite{touvron2021training}.} The experiment settings of dTPS-DeiT follows dynamicViT\footnote{ \url{https://github.com/raoyongming/DynamicViT}} except for basic learning rate is set to $\frac{batch size}{1024}\times 2.5\times10^{-4}$ and no stage of fixing backbone weights. The experiment settings of eTPS-DeiT follow EViT\footnote{\url{https://github.com/youweiliang/evit}}. The pruning settings include combinations of three multi-layer pruning settings: $prune\_locs \in \{[4,7,10],[3,5,7,9],[4,6,8,10]\}$, and two token keeping ratios :$ \rho \in \{0.5,0.7\}$. The token keeping ratio remains the same in all pruning stages.
    
    
    \textbf{TPS on LV-ViT~\cite{jiang2021all}.} The experiments of dTPS and eTPS  on LV-ViT~\cite{jiang2021all} follow the same training settings of dTPS and eTPS on DeiT, except for the basic learning rate is set to $\frac{batch size}{1024}\times 1.0\times10^{-4}$ for the stable convergence. For LV-ViT-T, the pruning settings include combinations of three multi-layer pruning settings: $prune\_locs \in \{[4,7,10],[3,5,7,9],[4,6,8,10]\}$, and two token keeping ratios :$ \rho \in \{0.5,0.7\}$. For LV-ViT-S, the pruning settings include combinations of three multi-layer pruning settings: $prune\_locs \in \{[5,9,13],[3,6,9,12],[4,7,10,13]\}$, and two token keeping ratios :$ \rho \in \{0.5,0.7\}$.
    
    \textbf{TPS on PS-ViT~\cite{yue2021vision}.} The experiments of dTPS on PS-ViT~\cite{yue2021vision} follow the same training settings as ATS~\cite{fayyazadaptive} on PS-ViT~\cite{yue2021vision}. The basic learning rate is set to $\frac{batch size}{768}\times 5.0\times10^{-4}$, $prune\_locs$ is set to [3,6,9] and token keeping ratio $\rho$ is 0.5. 
    
    \textbf{TPS on PVT~\cite{wang2021pyramid}.} The pruning stages include stage 2, stage 3, and stage 4. The token keeping ratio for all dTPS modules is set to 0.7. Basic learning rate is set to $\frac{batch size}{1024}\times 2.5\times10^{-4}$.
    
    \textbf{TPS on CvT~\cite{wu2021cvt}.} The basic learning rate is set to $\frac{batch size}{1024}\times 5.0\times10^{-5}$. The dTPS modules are only inserted into stage 3, and the pruning locations include [3,6,9] for CvT-13,[5,9,13] for CvT-21. The token keeping ratio for all TPS modules is set to 0.5. 
    
\subsection{iNaturalist 2019 Classification}
    \textbf{TPS on DeiT~\cite{touvron2021training}.} 
    For the experiment on iNaturalist 2019 Classification~\cite{fgvc6}, we re-train DeiT and fine-tune the model with dynamicViT or dTPS applied.
    
    In the training step, we initialize DeiT~\cite{touvron2021training} with weights of ImagetNet1K pre-trained model and re-train them for $300$ epochs. The basic learning rate is set to $\frac{batch size}{1024}\times10^{-3}$. The other settings follow DeiT~\cite{touvron2021training}. 
    
    In the fine-tuning step, we initialize dynamicViT-DeiT and dTPS-DeiT with weights from the last step and fine-tune them for 30 epochs with the same pruning setup. The token pruning location is set to [4,7,10], and the token keeping ratio is 0.5. We follow the same fine-tuning settings as the experiments on ImageNet1K, except for no distillation loss.

\section{TPS on Larger Models and with Larger Input Size}
We conduct experiments of TPS on DeiT-B as shown in \cref{DeiT_base} to demonstrate it is compatible with large models. We also prove TPS can perform well with larger input size such as $384 \times 384$, as shown in \cref{DeiT_384}.

\section{TPS under Different Keep Ratios}
\begin{figure}
    \centering
    \includegraphics[width=0.75\linewidth]{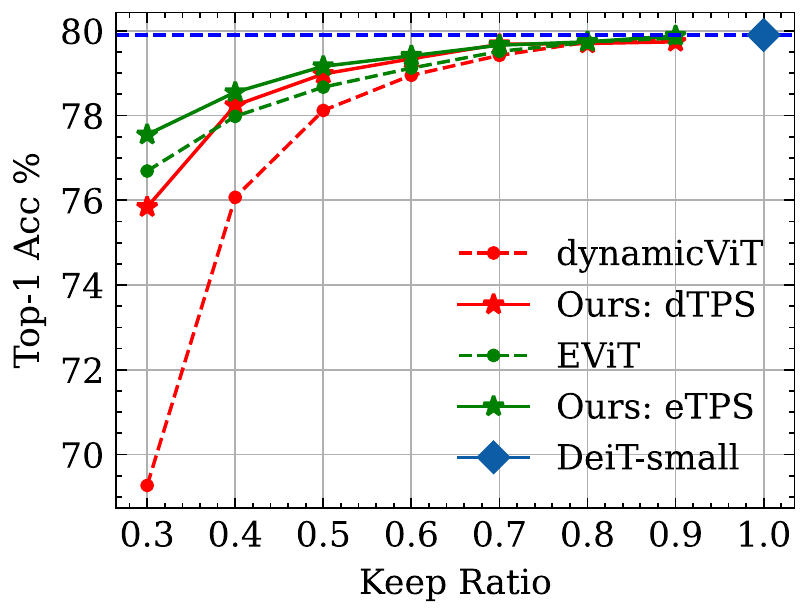}
    \caption{TPS under different keep ratios on DeiT-S.}
    \label{keep_ratio}
\end{figure}

Experiments of TPS under different keep ratios are shown in \cref{keep_ratio}.

\section{More Ablations About TPS Design}
More matching \& fusing methods are shown in \cref{ablation} as ablations about our TPS design. \cref{matching} indicates that TPS performance improvement benefits from compressing pruned tokens' information while unmatched reserved tokens remain unchanged. Token scoring can be proved necessary for squeezing under random token division meets a significant drop as shown in \cref{fusing} and robustness analysis in the main paper: Sec.~4.3.

\clearpage
{\small
\bibliographystyle{ieee_fullname}
\bibliography{egbib}
}

\end{document}